\documentclass[letterpaper, 10 pt, conference]{ieeeconf}  % Comment this line out if you need a4paper
\IEEEoverridecommandlockouts                              % This command is only needed if you want to use the \thanks command
\usepackage{setspace}
\usepackage{multirow}
\usepackage{graphicx}  % for \rotatebox
\usepackage{pifont}
\usepackage{tabularx, booktabs, adjustbox}
\usepackage{amssymb, amsfonts, bbold}
\usepackage{amsmath}
\usepackage{mathtools}
\usepackage{url}
\usepackage{cite}
\usepackage{color}
\usepackage[para]{footmisc} % Put footnote side by side, see here https://tex.stackexchange.com/a/28737

\usepackage{subcaption}
\usepackage{xcolor}

\makeatletter
\let\NAT@parse\undefined
\makeatother
\usepackage{hyperref}
\hypersetup{
	colorlinks=true,
	linkcolor=blue,
	filecolor=blue,      
	urlcolor=blue,
	citecolor=blue,
}
\usepackage{fancyhdr}
\fancypagestyle{withfooter}{
  
  \fancyhead[L]{}
  \fancyhead[R]{}
  \fancyfoot[C]{\footnotesize Presented at the 2025 IEEE ICRA Workshop on Field Robotics}
}
\def\R{\mathbb{R}}

\def\numTasks{n}
\def\numSamples{m}

\def\InputVec{\mathbf{x}}
\def\InputSet{\mathbf{X}}

\def\OutputVal{y}
\def\OutputVec{\mathbf{y}}
\def\OutputSet{\mathbf{Y}}

\def\StateVal{\theta}
\def\StateVec{\boldsymbol{\StateVal}}
\def\StateVecCross{\StateVec_c}
\def\StateVecSpace{\StateVec_s}

\def\Sampler{SAS}
\def\Analyzer{Lab}
\def\PrevLearner{STGP}
\def\LearnerName{MTGP}

\def\CovMatSpace{\mathbf{K}_s}
\def\CovMatCorr{\mathbf{K}_c}
\def\FullCovMat{\mathbf{K}}

\def\LowerMat{\mathbf{L}}

\title{\LARGE \bf
Towards Autonomous In-situ Soil Sampling and Mapping in Large-Scale Agricultural Environments
}

\author{Thien Hoang Nguyen, Erik Muller, Michael Rubin, Xiaofei Wang, Fiorella Sibona,\\
Alex McBratney, and Salah Sukkarieh % <-this % stops a space
\thanks{This work is supported by the Vonwiller Foundation and the University of Sydney's Digital Sciences Initiative (Corresponding author: Thien Hoang Nguyen). The authors are with the Australian Centre for Robotics (ACFR), University of Sydney, NSW, AU.
{\tt\small \{thienhoang.nguyen, erik.muller, michael.rubin, xiaofei.wang, fiorella.sibona, alex.mcbratney, salah.sukkarieh\}@sydney.edu.au}}%
}
\begin{document}

\thispagestyle{withfooter}
\pagestyle{withfooter}
%%% TO BE REMOVED (number of pages for reference while writing)
% \thispagestyle{plain}
% \pagestyle{plain}

\maketitle

%%%%%%%%%%%%%%%%%%%%%%%%%%%%%%%%%%%%%%%%%%
%               ABSTRACT                 %
%%%%%%%%%%%%%%%%%%%%%%%%%%%%%%%%%%%%%%%%%%
\begin{abstract}
Traditional soil sampling and analysis methods are labor-intensive, time-consuming, and limited in spatial resolution, making them unsuitable for large-scale precision agriculture. To address these limitations, we present a robotic solution for real-time sampling, analysis and mapping of key soil properties.
Our system consists of two main sub-systems: a Sample Acquisition System (\Sampler) for precise, automated in-field soil sampling; and a Sample Analysis Lab (\Analyzer) for real-time soil property analysis. The system’s performance was validated through extensive field trials at a large-scale Australian farm. Experimental results show that the \Sampler\ can consistently acquire soil samples with a mass of 50g at a depth of 200mm, while the \Analyzer\ can process each sample within 10 minutes to accurately measure pH and macronutrients. These results demonstrate the potential of the system to provide farmers with timely, data-driven insights for more efficient and sustainable soil management and fertilizer application.
\end{abstract}

%%%%%%%%%%%%%%%%%%%%%%%%%%%%%%%%%%%%%%%%%%
%                SECTION                 %
%%%%%%%%%%%%%%%%%%%%%%%%%%%%%%%%%%%%%%%%%%
\section{INTRODUCTION} \label{sec:Intro}

Achieving sustainable agricultural resource management requires accurate, high-resolution, and up-to-date data on soil properties such as pH and macronutrients \cite{mcfadden2023precision,dattatreya2023conventional}. However, conventional soil sampling and testing methods fail to address this need at scale. A typical soil sampling and analysis procedure involves multiple manual steps, including sample collection, sending the samples to a laboratory facility, chemical treatment, filtration, and analysis. The sample collection step demands significant effort from trained personnel, who must perform sampling, store samples appropriately, and deliver them promptly to ensure accurate analysis results \cite{rayment2011soil}. This process is not only time-consuming but also physically demanding, which becomes even more challenging as the number of samples increases. Furthermore, the results from laboratory facilities often take weeks to return, which further delays decision-making and makes these methods unsuitable for real-time, large-scale agricultural applications. Finally, conventional soil analysis methods yield discrete measurements of soil properties and nutrient levels at isolated sampling locations, which may limit farmers' ability to optimize resource use and to enhance productivity. In contrast, dense, spatially correlated data can provide a comprehensive understanding of the complex relationships between critical factors such as pH, macronutrient levels, and crop health. A more automated and reliable solution, designed for large-scale operation, would allow such granular data to be obtained more effectively. 

Robotic solutions enable real-time, in-field assessments, reducing reliance on centralized laboratories and accelerating data collection for applications in agriculture, environmental monitoring, and land management \cite{Nick2024ICRA}. Thus, there is a growing interest in developing robotic systems that can offer a transformative solution to the aforementioned challenges by automating sampling, preparation, and testing processes \cite{botta2022review,harun2023robotic}. The crux of a robotic solution is its ability to measure the target soil properties automatically, reliably, and in situ.
Unlike traditional laboratory-based soil testing methods, automation can reduce human labor, enhance efficiency, and improve the consistency and scalability of soil analysis. By providing rapidly acquired, high-frequency, and spatially dense data, robotic systems can support sustainable practices, minimize waste, and maximize productivity.

% \begin{figure}[t]
% \centering
%     \includegraphics[width=\linewidth]{Figures/PlatformOverviewTwoRows.png}
%     \caption{3D model and the real system in the field: (a-b) \Sampler, (c-d) \Analyzer.}
% \label{fig:PlatformOverview}
% \end{figure}

\begin{figure}[t]
\captionsetup[subfigure]{justification=centering}
\centering
    \begin{subfigure}{.445\columnwidth}
        \centering
        \includegraphics[width=\linewidth]{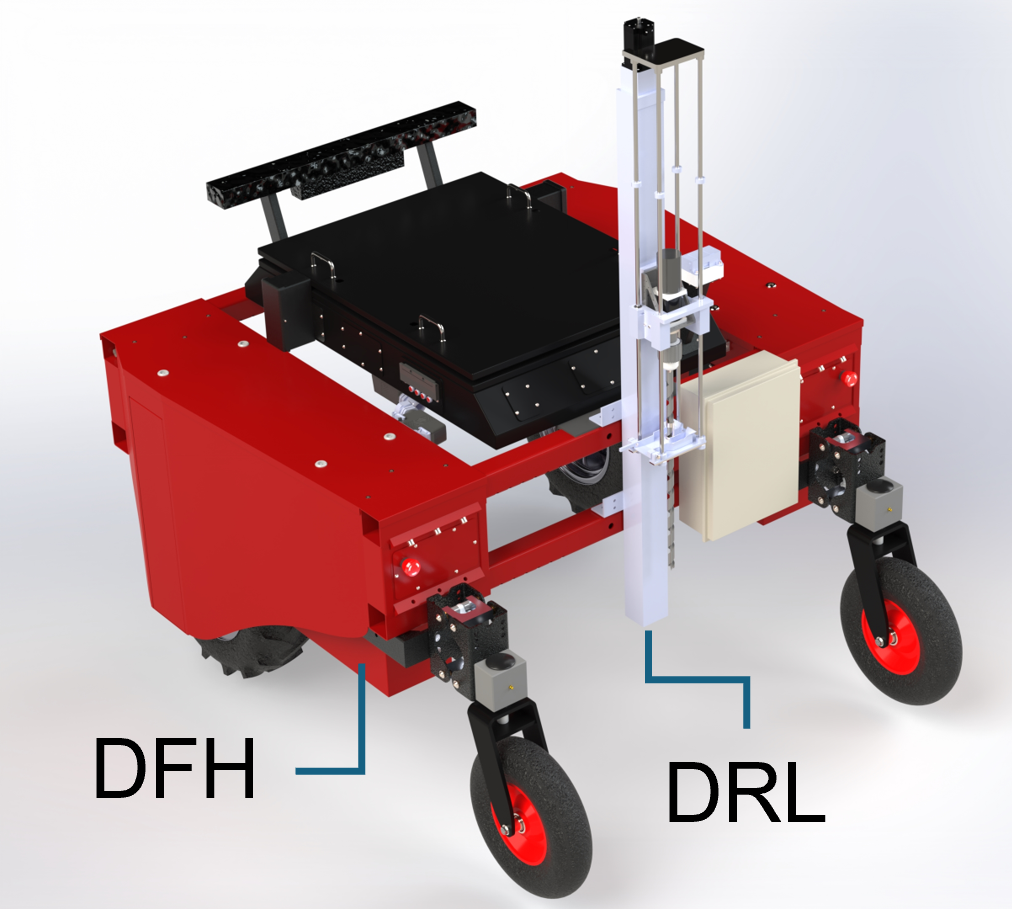}
        \caption{3D model of the \Sampler}
        \label{fig:PlatformOverview_a}
    \end{subfigure}
    \begin{subfigure}{.53\columnwidth}
        \centering
        \includegraphics[width=\linewidth]{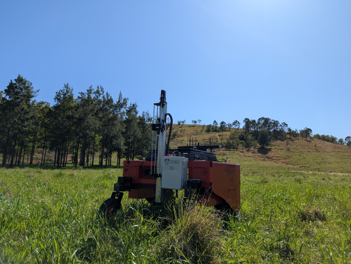}
        \caption{Real \Sampler\ sub-system}
        \label{fig:PlatformOverview_b}
    \end{subfigure}
    \begin{subfigure}{.445\columnwidth}
        \centering
        \includegraphics[width=\linewidth]{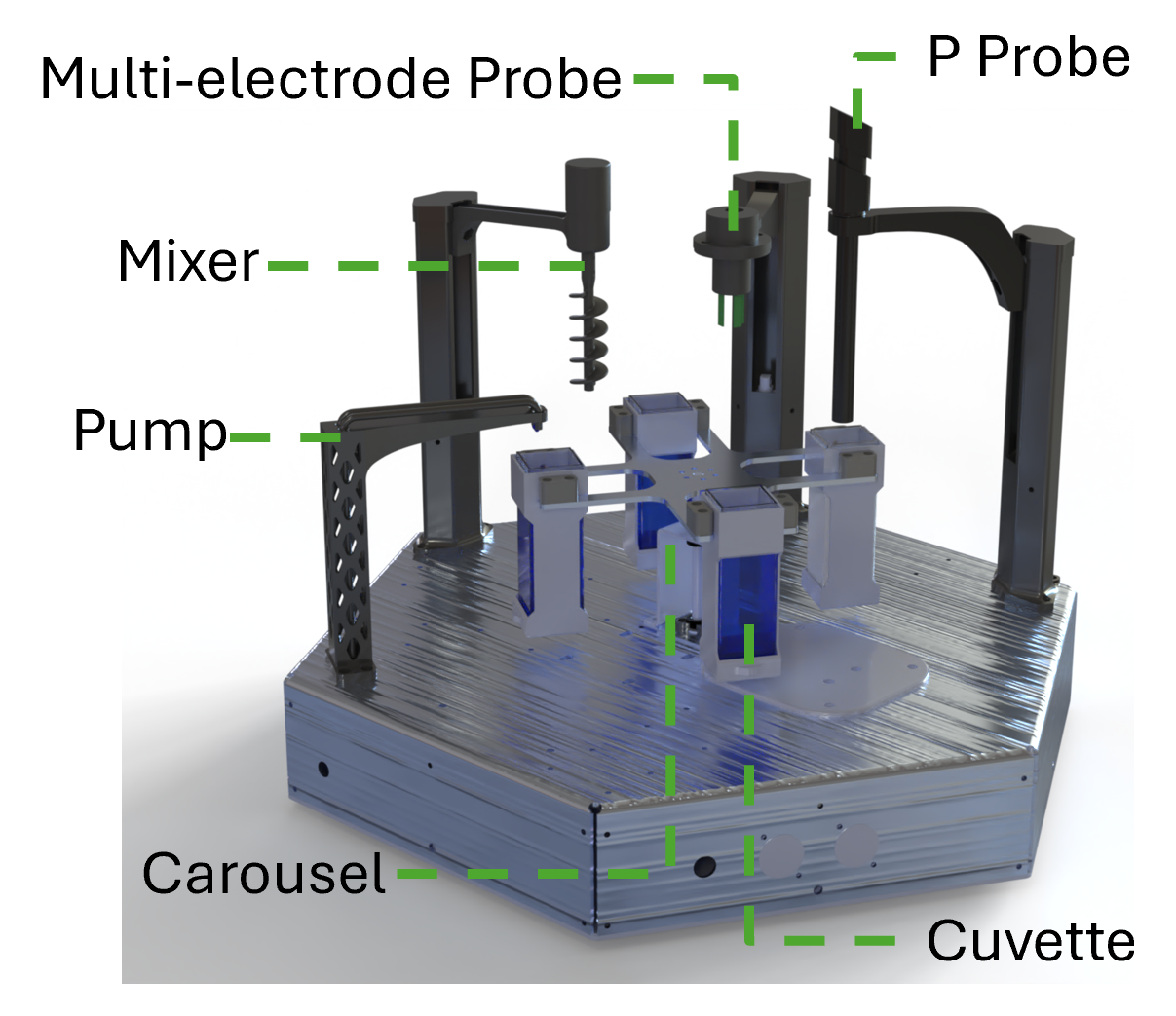}
        \caption{3D model of the \Analyzer}
        \label{fig:PlatformOverview_c}
    \end{subfigure}
    \begin{subfigure}{.53\columnwidth}
        \centering
        \includegraphics[width=\linewidth]{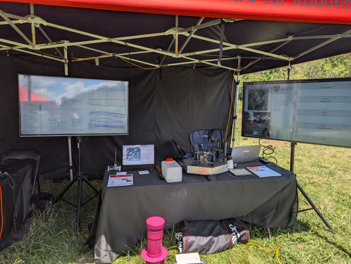}
        \caption{Real \Analyzer\ sub-system}
        \label{fig:PlatformOverview_d}
    \end{subfigure}
    \caption{3D model and the real-world system in the field.}
    \label{fig:PlatformOverview}
\end{figure}

In this paper, we present a novel robotic system capable of autonomous real-time sampling and mapping of key soil properties (Fig. \ref{fig:PlatformOverview}). Our main contributions include:
\begin{itemize}
    \item Sample Acquisition System (\Sampler): A mobile robot equipped with an advanced sampling mechanism that can autonomously navigate to target locations, drill to precise depths, collect soil samples of specified mass, and return the samples to the base station;
    \item Sample Analysis Lab (\Analyzer): A mobile unit for on-site measurements of targeted soil properties, including pH and concentrations of Nitrogen (N), Phosphorus (P), and Potassium (K), featuring a semi-autonomous workflow to handle soil sample processing, data analysis, and environmental modeling tasks;
    \item A modular system architecture comprising the \Sampler\ and \Analyzer\ subsystems, where the \Sampler\ integrates a mobile robot and a drilling unit, both of which can be independently adapted or replaced to suit different terrains, soil types, and mission requirements;
    \item Extensive real-life experiments on a 50,000 m$^2$ pasture farm validate the performance and robustness of our system under real-world farming conditions.
\end{itemize}

%%%%%%%%%%%%%%%%%%%%%%%%%%%%%%%%%%%%%%%%%%
%                SECTION                 %
%%%%%%%%%%%%%%%%%%%%%%%%%%%%%%%%%%%%%%%%%%
\section{LITERATURE REVIEW} \label{sec:LitReview}

Although various robotic solutions for in-field soil sampling or analysis have been put forth, most remain at the prototyping stage \cite{harun2023robotic,botta2022review}. As such, we focus on systems that have been rigorously field-tested and validated, and examine the challenges that hinder their broader adoption. The automated on-the-go soil nitrate mapping system (SNMS) \cite{sibley2008development} and Agrobot Lala \cite{kitic2022agrobot} are autonomous robotic systems designed to collect soil samples and analyze nitrate (NO$_3^-$) levels directly in the field. Both systems utilize specialized actuators for soil collection and Ion-Selective Electrode (ISE) sensors for direct Nitrogen measurement. However, they differ in core functionality for the soil analysis task: while SNMS is entirely manual and lacks sample handling or processing capabilities, Agrobot incorporates both features. Nonetheless, Agrobot has several notable limitations that constrain its broader applicability. One significant drawback is its focus on analyzing only nitrate levels, limiting its utility for comprehensive soil nutrient mapping that includes other macronutrients such as Phosphorus and Potassium. Additionally, while the system provides real-time analysis, the process requires approximately 30 minutes per sample, which could hinder scalability for large fields requiring high-density sampling. 

The Soil2data project \cite{hinck2022prototypes4soil2data} is another innovative approach to automated, on-the-go soil testing for agricultural applications. Integrated with the Bonirob robot \cite{ruckelshausen2009bonirob}, the Soil2data system is capable of autonomous navigation, sampling, analysis, and even fertilizer recommendation. The system employs a 16 mm auger, which can drill up to 30 cm to collect representative soil samples for nutrient analysis. The sensor module, equipped with ion-sensitive field-effect transistor (ISFET) sensors, provides real-time measurements of key soil parameters such as pH and NPK, with concentration levels up to 100 mMol/l. Soil processing and analysis are completed in approximately 15 minutes, with the ISFET sensors stabilizing within 100 seconds. Once testing is finished, the soil samples are left in the field, minimizing the need for transport and laboratory work. 

The NASA's Curiosity rover, launched in 2012, features an advanced on-board laboratory called the Sample Analysis at Mars (SAM)  \cite{grotzinger2012mars}. This system is designed to analyze Martian soil and rock samples for the presence of key elements like carbon, oxygen, hydrogen, and nitrogen, which are essential for understanding Mars' potential to support life. The SAM system is equipped with a sample manipulation system that transports powdered samples from the rover’s drill to the analysis chambers.
This combination of tools allows the Curiosity rover to conduct comprehensive in-situ analyses of Martian soil.

Even though the Soil2data and Curiosity systems represent significant advancements in automated soil analysis, several challenges remain in terms of reproducibility and widespread adoption. First, to employ the ISFET multi-sensor module, which is custom-made and not commercially available, Soil2data is meticulously engineered to ensure accuracy and reliability, addressing potential issues such as dust, light interference, moisture, mechanical stress, corrosion, and cross-sensor interference \cite{riedel2024concept}. Second, as the analysis occurs onboard the robot, a considerable amount of chemicals (e.g., extraction reagents, calibration solutions, and deionized water) must be carried for processing. Third, while the maximum number of samples the system can process before needing a chemical refill has not been reported, it is proportional to the amount of chemicals onboard the robot. Hence, the more samples it wishes to process, the more chemicals it needs to carry. This mass burden further reduces the robot’s run time and overall mission duration, requiring it to return to base for replenishment.
Similarly, the Curiosity is a highly specialized platform with substantial development and operational costs, making it impractical for replication in research and commercial applications on Earth.

Compared to existing approaches, our system offers several distinct advantages by separating the mobile robot for sampling and the stationary lab for processing and analysis. Firstly, our system is a more cost-effective and adaptable solution for in-field soil sampling and analysis. In our design, the autonomous robot focuses exclusively on in-field soil sampling, while the nutrient analysis is performed at a centralized base station. This separation reduces the design complexity, weight constraints, and power demands on the mobile robot, enhancing its operational efficiency and mission duration. Secondly, placing the sensitive ISFET sensors at the base station allows them to be protected from disturbances caused by the robot’s operation, such as vibrations, dust, moisture and temperature fluctuations. A fully integrated system would introduce substantial mechatronic design challenges to prevent the above factors from compromising measurement accuracy, which our system inherently avoids. This modular architecture also improves upgradability and repairability, as the analysis system can be easily upgraded to include other elements beyond NPK and the sensors can be easily replaced when they fail, all without disassembling or grounding the robot. This decoupled design provides a cost-effective and robust advantage for our system, which is highly desirable to agricultural applications.

%%%%%%%%%%%%%%%%%%%%%%%%%%%%%%%%%%%%%%%%%%
%                SECTION                 %
%%%%%%%%%%%%%%%%%%%%%%%%%%%%%%%%%%%%%%%%%%
\section{SYSTEM DESIGN} \label{sec:Systems}

%%%%%%%%%%%%%%%%%%%%%%%%%%%%%%%%%%%%%%%%%%
%============ [ Subsection ] ============%
%%%%%%%%%%%%%%%%%%%%%%%%%%%%%%%%%%%%%%%%%%
\subsection{System Overview}

\begin{figure*}[t]
\centering
    \includegraphics[width=0.9\linewidth]{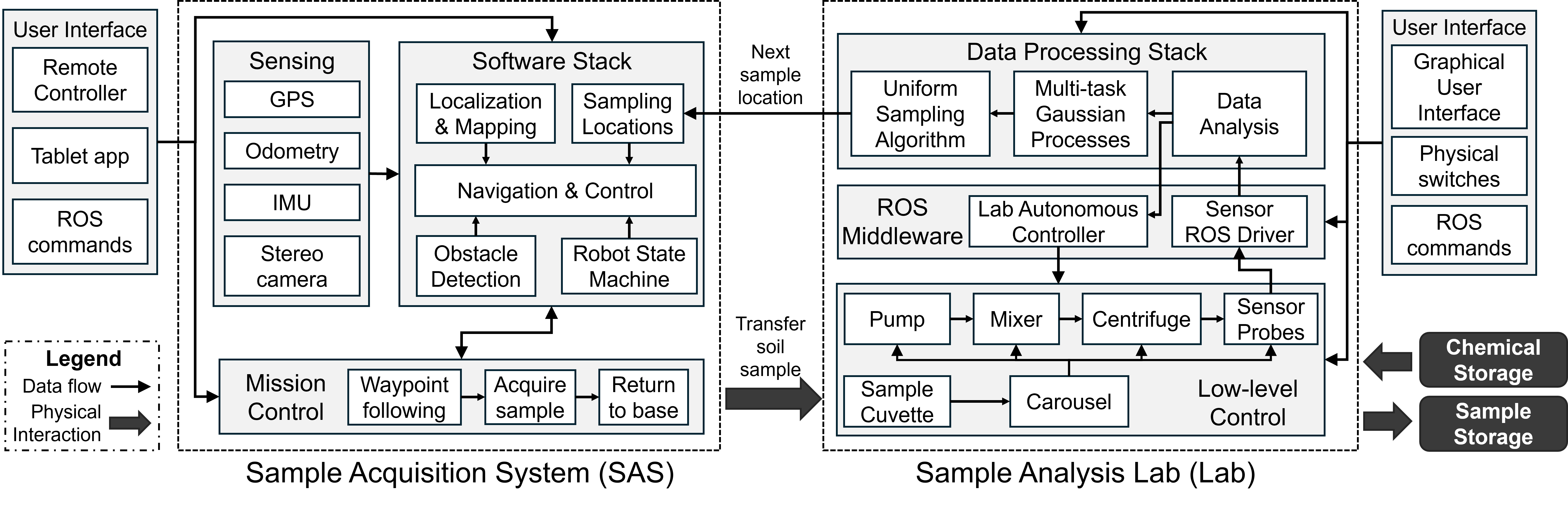}
    \caption{Overview of the proposed system, which consists of two main sub-systems: the \Sampler\ and \Analyzer.}
\label{fig:SystemOverview}
\end{figure*}

\begin{figure*}[t]
\centering
    \includegraphics[width=1\linewidth]{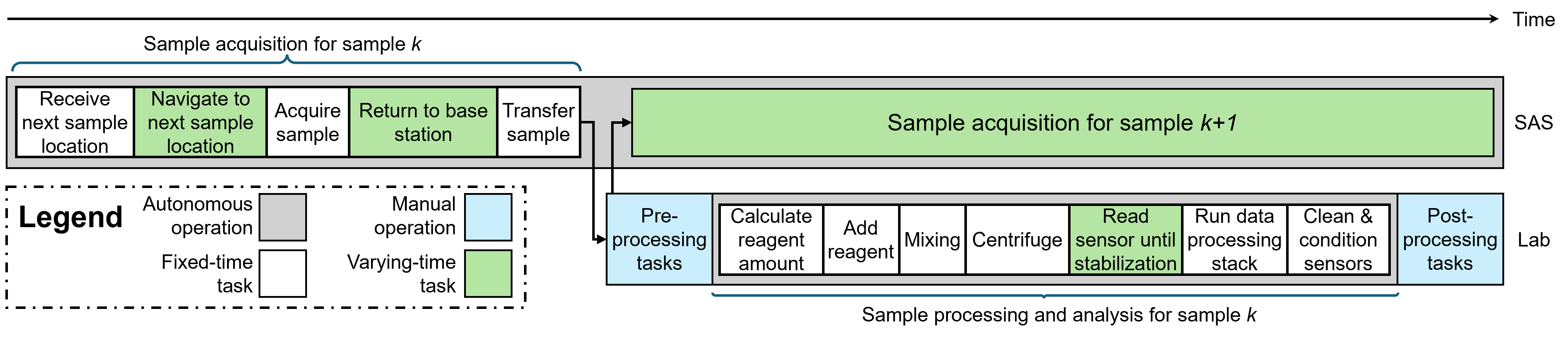}
    \caption{High-level operational flowchart of the proposed system.}
\label{fig:SWPipeline}
\end{figure*}

As can be seen in Fig. \ref{fig:SystemOverview}, our system comprises two main physical components: the \Sampler\ and the \Analyzer. In a nutshell, the \Sampler\ is responsible for soil sample acquisition and metadata collection, while the \Analyzer\ is responsible for processing the sample. During a typical operation, the \Sampler\ autonomously navigates to designated locations using GPS, odometry, and localization systems, performs drilling to reach specified depths, collects soil samples, and returns to the base station. Here, the soil sample is manually transferred to the \Analyzer\ for handling, processing and analysis. Once the sample is received by the \Analyzer, \Sampler\ automatically continues to the next sample location. Simultaneously, the \Analyzer\ executes the sample processing and analysis workflow to process the sample and measure the targeted soil properties. To perform the aforementioned tasks, the \Sampler\ and \Analyzer\ sub-systems are designed to meet several critical functional and quality control requirements, ensuring their effectiveness and reliability in the real world, which are listed in Table \ref{table:design_requirements}.

\begin{table}[ht]
\centering
\begin{tabular}{p{0.95\linewidth}}
\toprule
\textbf{Design requirements for \Sampler} \\ 
\hline         
    Collect 30-50g samples from 0-200mm depth with $\leq 10\%$ contaminants\\
    Perform a complete fully autonomous sample acquisition without human intervention \\
    Log essential operational data for post-analysis (e.g., GPS locations, sample depth and weight, travel time).\\
%%%%%%%%%%%%%%%%%%%%%%%%%%%%%%%%%%%%%%%%%%%%%%%%%%%%%%%%%%%%%%%%%
\toprule
\textbf{Design requirements for \Analyzer} \\ 
\hline
    Accurately measure pH, N, P, and K with a fixed soil-to-reagent of ${r=1:3}$\\
    Perform a complete sample processing and analysis process without human intervention\\
    Sensors must be shielded to reduce light, moisture, vibration and electromagnetic interferences \\
    % Auto-adjust the amount of reagent added based on sample mass to maintain a fixed soil-to-reagent ratio. \\
    Record operational data for post-analysis (e.g. sensors' data, sample weight, processing and analysis time).\\
\bottomrule
\end{tabular}
\caption{Key design requirements for the proposed system.}
\label{table:design_requirements}
\end{table}

%%%%%%%%%%%%%%%%%%%%%%%%%%%%%%%%%%%%%%%%%%
%============ [ Subsection ] ============%
%%%%%%%%%%%%%%%%%%%%%%%%%%%%%%%%%%%%%%%%%%
\subsection{Sample Acquisition System (\Sampler)}

The \Sampler\ comprises two main components: the Digital Farmhand (DFH) robot, and the Dirt Requisition and Lifting (DRL) module which is attached to the DFH. The DFH is a versatile, solar-electric ground robot designed to perform automated agriculture tasks such as weeding and crop assessment. The DRL module has been developed to equip DFH with the capability to reliably collect soil samples of known mass and depth.
Fig. \ref{fig:SWPipeline} illustrates the high-level flowchart of the full system's operation and the step-by-step flowchart of the \Sampler's operation, respectively.

The most critical considerations during the design phase for the DRL are the sampling depth and sample mass. The sampling depth is measured by equipping the DRL with a 1D Lidar module to obtain the height of the drill relative to the ground, then the linear actuator will adjust the deployment length accordingly. By design, the maximum depth that the drill can reach is 243mm, thus allowing a 21.5\% extension margin for the system to collect samples at 200mm below the surface. This extension margin is necessary since in real-world conditions, the terrain is often not completely flat and adjustment might be needed to ensure the drill reaches the target depth below topsoil.
Additionally, the sample mass is calculated as:
\begin{equation}
m_s = \rho \times V = \rho \times \pi \times L \times (d/2)^2,
\end{equation}
where $m_s$ is the mass of the soil sample (g), $\rho$ is the bulk density of the sample (g/mm$^3$), $V$ is the sample volume (mm$^3$), $L$ is the depth of the sample (mm), and $d$ is the auger’s diameter (mm). Note that the design is based on the hypothesis that by extracting a sample of fixed volume, the sample’s mass can be reliably estimated under the assumption of a uniform bulk density. As a result, augers with different diameters $d$ are needed to collect the same amount of soil $m_s$ at areas with different bulk densities $\rho$.

%%%%%%%%%%%%%%%%%%%%%%%%%%%%%%%%%%%%%%%%%%
%============ [ Subsection ] ============%
%%%%%%%%%%%%%%%%%%%%%%%%%%%%%%%%%%%%%%%%%%
\subsection{Sample Analysis Lab (\Analyzer)}

The role of the \Analyzer\ is processing and analyzing the soil samples obtained by the \Sampler\ through a semi-automated workflow. 
The \Analyzer\ is designed based on the sample preparation, processing, and analysis pipeline for in-field measurement of pH and NPK, as outlined in \cite{najdenko2023development}. As the ISFET measurement module in \cite{najdenko2023development} is not commercially available, we adopted the pipeline with necessary modifications for off-the-shelf sensors and components. Fig. \ref{fig:SWPipeline} shows the flowchart of the \Analyzer\ and how it fits in the overall system. Pre-processing and post-processing tasks include removing stones, roots, or other debris from the sample and cleaning the cuvettes for the next processing cycle, respectively.

Our system integrates pH, N, and K ISFET sensors from Microsens SA\footnote{\url{https://microsens.ch/products/ISFET.htm}}, and P sensor from EDT directION\footnote{\url{https://edt.co.uk/product/phosphate-electrode-kit}}. The reagent is 0.01 M CaCl2, which is an effective option for N and K analysis \cite{najdenko2023development}. Since the sensors from Microsens SA can work simultaneously using a shared reference electrode, a single-reference multi-electrode probe which integrates the pH, N, and K ISFET sensors was developed. Thanks to this innovation, we can submerge a single reference electrode and multiple sensors in the same solution and simultaneously read out measurements from all of the sensors. 
% Since the outputs of the sensors are provided in millivolts (mV), the ion concentration values in ppm are then derived through the automatic data analysis process as follows. Let $S$ (measured in mV) and $C$ (measured in ppm) be the stable sensor output and ion concentration, respectively. Within the measurement range, the relationship between $S$ and $C$ (in log\textsubscript{10} steps) is linear and can be written as:
% \begin{equation}\label{eq:ISFETCalib}
% \log_{10}(C / r) = A S + B \quad \textrm{or} \quad C = r 10^{A S + B},
% \end{equation}
% where \(A\) and \(B\) are the linear fit parameters that can be determined using 3-point calibration, and $r$ is the soil-to-reagent ratio, i.e. the ratio between the added reagent's volume (in mL) and the soil sample's mass (in grams), also referred to as the dilution factor.

%%%%%%%%%%%%%%%%%%%%%%%%%%%%%%%%%%%%%%%%%%
%============ [ Subsection ] ============%
%%%%%%%%%%%%%%%%%%%%%%%%%%%%%%%%%%%%%%%%%%
\subsection{Soil Properties Mapping using MTGP}

Gaussian Processes (GPs) are widely used for modeling spatial correlations of environment properties with sparse data \cite{williams2006gaussian}. For soil mapping applications, GPs offer several key advantages. First, they enable predictions of soil properties across different spatial scales while providing corresponding uncertainty estimates. Second, GPs yield well-defined confidence intervals, which are crucial for soil scientists to assess model reliability. In this work, we adopt the Multi-task Gaussian Processes (MTGP) framework proposed in \cite{bonilla2007multi} to simultaneously model multiple correlated soil properties. In contrast to using separate Single-task Gaussian Processes (\PrevLearner) for each variable, MTGP can learn correlations between different variables, even when sampled at varying frequencies or with training data available over different intervals. Moreover, it allows seamless integration of prior knowledge about variable relationships to further enhance the predictive accuracy.
This approach is particularly suited for field robotics applications, where sample collection is costly, and measurements are often sparse or unevenly distributed across tasks.

%%%%%%%%%%%%%%%%%%%%%%%%%%%%%%%%%%%%%%%%%%
%=========== [ Subsubsection ] ==========%
%%%%%%%%%%%%%%%%%%%%%%%%%%%%%%%%%%%%%%%%%%
\subsubsection{Covariance Function}

MTGP requires a covariance (or kernel) function to model the similarity between any two points in space. In the simplest case with two quantities of interest, MTGP uses two independent covariance functions to represent the correlation between tasks (\(k_c\)) and spatial covariance within a task (\(k_s\)) \cite{durichen2014multitask}:
\begin{equation}
k_{\text{MTGP}}(\InputVec,\InputVec') = k_c \times k_s(\InputVec,\InputVec'),
\end{equation}
where $\InputVec$ and $\InputVec'$ are the sample locations. It is clear that the cross-correlation function \(k_c\) is independent of the sampling locations $\InputVec$ and $\InputVec'$. 

Let $\numTasks$ be the number of quantities of interest ($\numTasks = 4$ for our target application) and $\numSamples$ be the number of observations. At a sample location ${\InputVec \in \R^2}$, a measurement vector
${\OutputVec(\InputVec) = [\OutputVal_1(\InputVec),\dots,\OutputVal_n(\InputVec)]^\top}$, which consists of $\numTasks$ individual measurements for $\numTasks$ quantities, can be obtained. Denote ${\InputSet = \{\InputVec_j \mid j = 1,\dots,\numSamples\}}$ and ${\OutputSet = \{\OutputVec_j \mid j = 1,\dots,\numSamples\}}$ as the set of sampling locations and corresponding measurement vectors. The full covariance matrix \(\FullCovMat\) for all \(\numTasks\) tasks can be written as \cite{bonilla2007multi}:
\begin{equation}
\FullCovMat(\InputSet, \mathbf{l}, \StateVecCross, \StateVecSpace) 
= \CovMatCorr(\mathbf{l}, \StateVecCross) \otimes \CovMatSpace(\InputSet, \StateVecSpace)
\end{equation}
where \(\mathbf{l} = \{i \,|\, i = 1, \dots, \numTasks\}\), \(\otimes\) is the Kronecker product, \(\StateVecCross\) and \(\StateVecSpace\) are vectors containing all hyperparameters for \(\CovMatCorr\) and \(\CovMatSpace\), respectively. 
% This leads to a matrix of size \(mn \times mn\) for \(\FullCovMat\), as \(\CovMatCorr\) has a size of \(m \times m\), and \(\CovMatSpace\) of \(n \times n\).
Using the so-called free-form covariance matrix \cite{Nick2024ICRA,williams2006gaussian}, the inter-task correlation matrix is formulated as \(\CovMatCorr = \LowerMat \LowerMat^\top\), where $\LowerMat \in \R^{n\times n}$ is a lower triangular matrix.
In this manner, \(\StateVecCross\) will contain the $N = n(n+1)/2$ non-zero elements of $\LowerMat$ to be estimated. 
On the other hand, $\CovMatSpace$ is constructed as follows. Let \(K^s_{ij}\) be the $(i,j)$-th element of \(\CovMatSpace\). $\CovMatSpace$ is devised by first modeling the task-specific covariance functions ($K^s_{ii}$), then convoluting them to obtain the cross-task covariance functions ($K^s_{ij}$). This process, described in detail in \cite{melkumyan2011multi}, preserves the spatial characteristics of each task while ensuring that the resulting matrix remains positive semidefinite.

We use the Matérn 3/2 covariance function to model the spatial structure of each individual task. This kernel is widely adopted in geospatial modeling due to its ability to capture both smooth and moderately rough variations across space \cite{williams2006gaussian}. The convolution-based formulation allows us to derive cross-task spatial covariances analytically, using only the parameters of the task-specific kernels. This provides an elegant and computationally efficient way to encode spatial relationships between soil properties, even when the sampling is non-uniform or tasks are only partially observed.

\subsubsection{Learning and Prediction}

% The most effective way to determine the hyperparameters $\StateVec = [\StateVecCross^\top, \StateVecSpace^\top]^\top$ in MTGP is by learning them directly from the training set \( \mathbf{T} = \{\InputSet, \OutputSet\} \), which can be achieved by minimizing the negative log marginal likelihood (NLML) as described in \cite{durichen2014multitask}:
% \begin{equation} \label{eq:NLML}
% \text{NLML}
% = \frac{1}{2} \log |\FullCovMat| 
% + \frac{1}{2} \OutputSet^\top \FullCovMat^{-1} \OutputSet
% + \frac{n}{2} \log (2\pi).
% \end{equation}
% In essence, the first term in Equation \ref{eq:NLML} penalizes model complexity, while the second term penalizes low data likelihood. Once the hyperparameters have been estimated, predictions values for a test location \(\InputVec^*\) can be made by computing the conditional probability:
% \begin{equation}
% p(\OutputVec^*|\InputVec^*, \InputSet, \OutputSet) \sim \mathcal{N}(m(\OutputVec^*), \text{var}(\OutputVec^*))
% \end{equation}
% with a mean \( m(\OutputVec^*) \) and variance \( \text{var}(\OutputVec^*) \), with the mean function \( m(\InputVec) \) is commonly assumed to be zero.
% The correlation coefficient $r_{ij}$ between the $i$-th and $j$-th quantities is computed as \cite{Nick2024ICRA}:
% \begin{equation}
%     r_{ij} = \frac{K^c_{ij}}{\sqrt{K^c_{ii}}\sqrt{K^c_{jj}}}
% \end{equation}
% where \(K^c_{ij}\) is the $(i,j)$-th element of \(\CovMatCorr\). If $r^2_{ij}$ is closer to $0$, the quantities are more likely to be independent of each other. Conversely, the closer $r^2_{ij}$ to $1$, the more strongly correlated the quantities are.

The most effective way to determine the hyperparameters $\StateVec = [\StateVecCross^\top, \StateVecSpace^\top]^\top$ in MTGP is by learning them directly from the training data through marginal likelihood optimization, as described in \cite{durichen2014multitask}. This approach balances model complexity with data fit, ensuring generalization while accurately capturing spatial and inter-task structure. Once the model is trained, predictions at unseen locations are obtained using the conditional distribution of a Gaussian process, which yields both the estimated mean and the associated uncertainty for each soil property.

In addition to predictions, the MTGP model allows us to compute the correlation between different soil variables based on the learned task covariance matrix. The correlation coefficient $r_{ij}$ between the $i$-th and $j$-th quantities is computed as \cite{Nick2024ICRA}:
\begin{equation}
    r_{ij} = \frac{K^c_{ij}}{\sqrt{K^c_{ii}}\sqrt{K^c_{jj}}}
\end{equation}
where \(K^c_{ij}\) is the $(i,j)$-th element of \(\CovMatCorr\). If $r^2_{ij}$ is close to $0$, the quantities are likely independent, while values near $1$ indicate strong correlation.

%This also supports adaptive sampling, where future measurements can be directed toward regions of high uncertainty.

% Without loss of generality, the mean function \( m(\InputVec) \) is commonly assumed to be zero.
% Under this assumption, \( m(\OutputVec^*) \) and \( \text{var}(\OutputVec^*) \) are given by:
% \begin{align}
% m(\OutputVec^*) 
% &= \FullCovMat(\InputVec, \InputVec^*)^\top \FullCovMat(\InputVec, \InputVec)^{-1} \OutputVec, \\
% \text{var}(\OutputVec^*) 
% &= \FullCovMat(\InputVec^*, \InputVec^*) - \FullCovMat(\InputVec, \InputVec^*)^\top \FullCovMat(\InputVec, \InputVec)^{-1} \FullCovMat(\InputVec, \InputVec^*).
% \end{align}

%%%%%%%%%%%%%%%%%%%%%%%%%%%%%%%%%%%%%%%%%%
%                SECTION                 %
%%%%%%%%%%%%%%%%%%%%%%%%%%%%%%%%%%%%%%%%%%
\section{EXPERIMENTAL RESULTS} \label{sec:Results}

%%%%%%%%%%%%%%%%%%%%%%%%%%%%%%%%%%%%%%%%%%
%============ [ Subsection ] ============%
%%%%%%%%%%%%%%%%%%%%%%%%%%%%%%%%%%%%%%%%%%
\subsection{Real-world Experiments Setup}

% \begin{figure}[t]
% \centering
%     \includegraphics[width=\linewidth]{Figures/FieldTrialOverview.png}
%     \caption{a) Overview of the pasture farm in Allynbrook, NSW, Australia. b) Aerial view of a section of the experimental area.}
% \label{fig:FieldTrialOverview}
% \end{figure}

\begin{figure}[t]
\captionsetup[subfigure]{justification=centering}
\centering
    \begin{subfigure}{1\columnwidth}
        \centering
        \includegraphics[width=\linewidth]{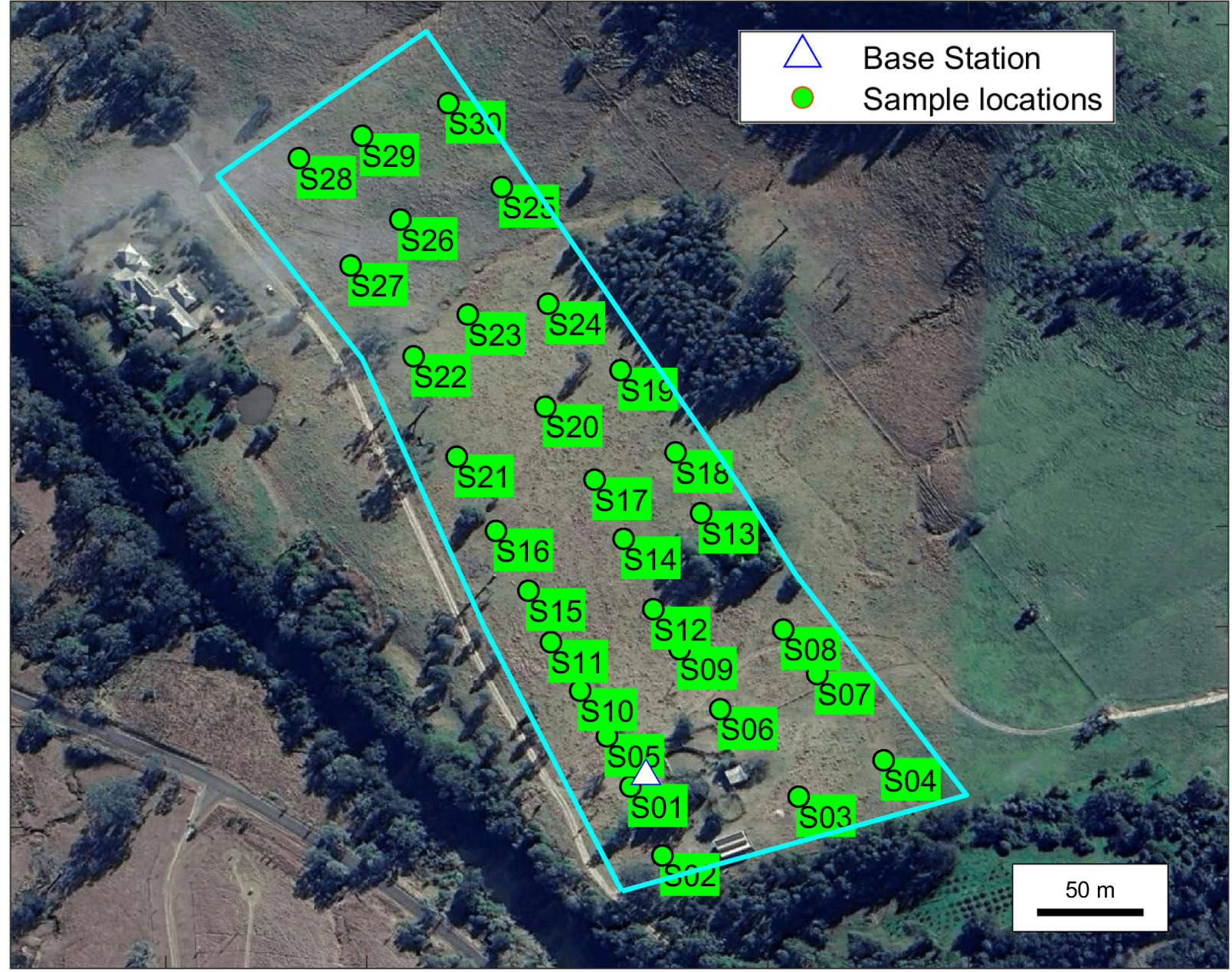}
        % \caption{}
        \label{fig:FieldTrialOverview_a}
    \end{subfigure}
    % \begin{subfigure}{.8\columnwidth}
    %     \centering
    %     \includegraphics[width=\linewidth]{Figures/FieldTrialOverview_b.png}
    %     \caption{}
    %     \label{fig:FieldTrialOverview_b}
    % \end{subfigure}
    \caption{Overview of the farm in Allynbrook, NSW, Australia.}
    \label{fig:FieldTrialOverview}
\end{figure}

\begin{figure}[t]
\captionsetup[subfigure]{justification=centering}
\centering
    \begin{subfigure}{.63\columnwidth}
        \centering
        \includegraphics[width=\linewidth]{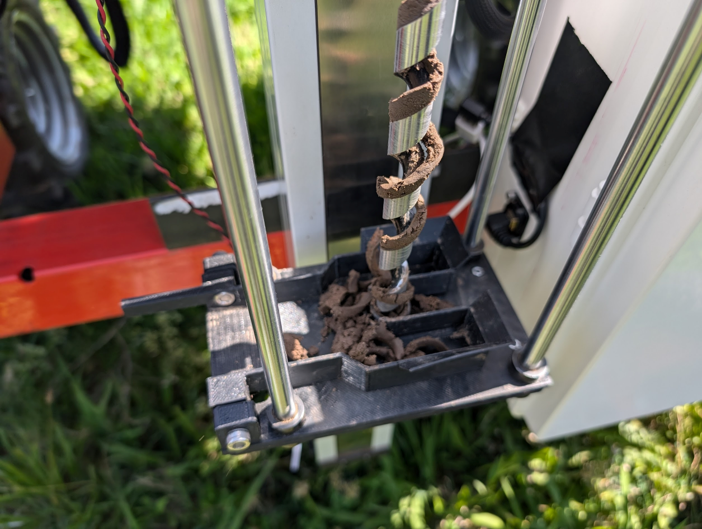}
        \caption{}
        \label{fig:ExpRealSample_a}
    \end{subfigure}
    \begin{subfigure}{.3414\columnwidth}
        \centering
        \includegraphics[width=\linewidth]{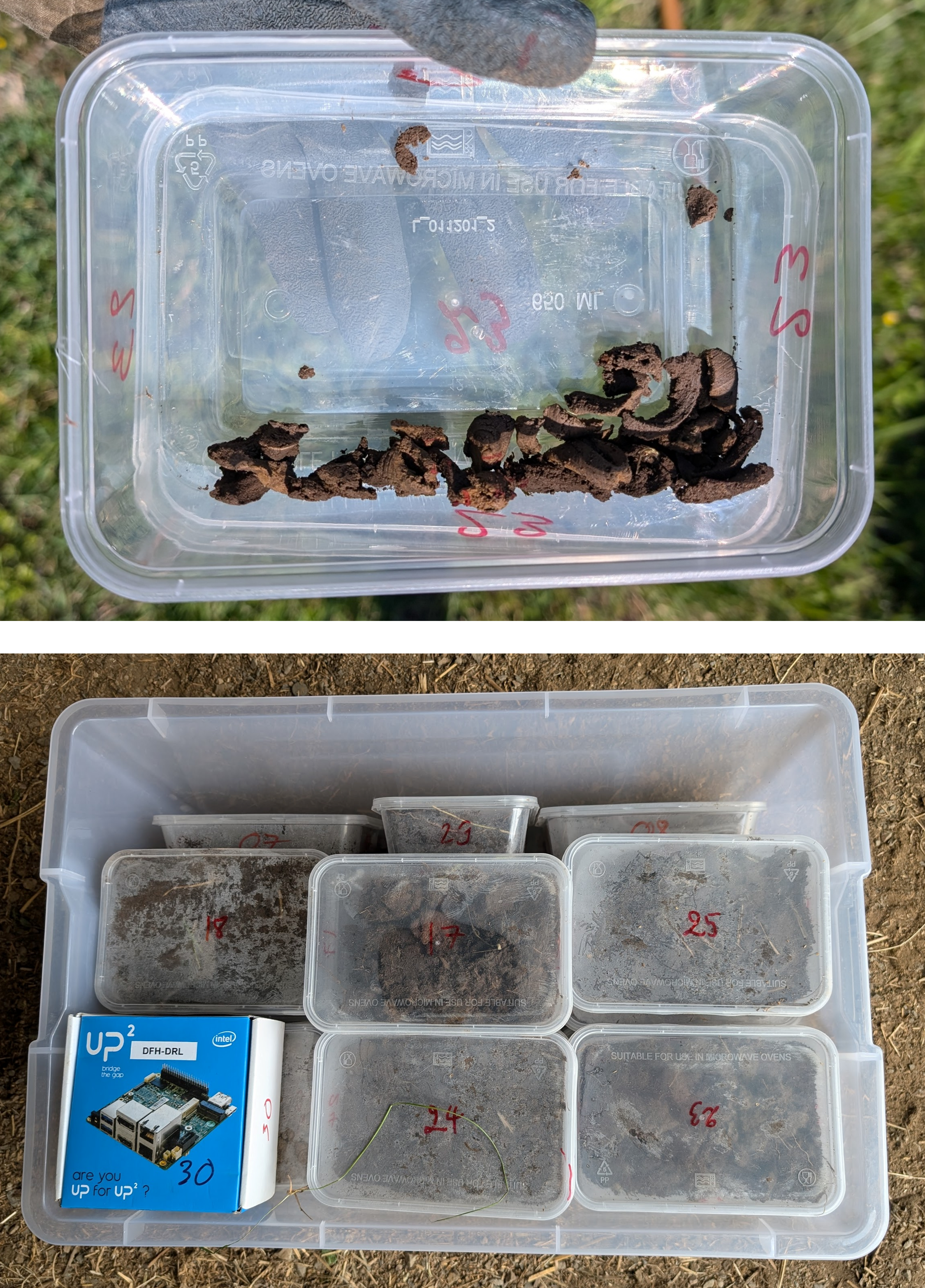}
        \caption{}
        \label{fig:ExpRealSample_b}
    \end{subfigure}
    \caption{(a) A soil sample acquired by the \Sampler. (b) A soil sample getting transferred to the \Analyzer\ and samples collected for ground truth analysis.}
    \label{fig:ExpRealSample}
\end{figure}

To validate our system in real agricultural conditions, we conduct extensive experiments at a 50,000 m$^2$ pasture farm in Allynbrook, NSW, Australia (Fig. \ref{fig:FieldTrialOverview}). The proposed system was deployed to collect, process, and analyze samples during a multi-day field trial. 
Before the field trial, the sample locations were chosen in a grid pattern that covers the field, with an average spacing of approximately 45 m between any two samples. This spacing has been determined in \cite{sibley2008development} to be a good choice of variogram for experimental sampling schemes, i.e. the range at which sampling would have to take place to make an accurate map of the Nitrogen variation across the field. Additionally, the sample locations need to avoid obstacles (e.g., trees and buildings), steep slopes and restricted regions on the farm. 
Based on these requirements, 30 sample locations were selected on Google Maps. 
At each sample location, an additional 0.5 kg of soil samples were manually collected and later sent to a commercial laboratory\footnote{\url{https://www.nutrientadvantage.com.au/}} in Victoria, Australia, to obtain ground truth values for NPK at a cost of $30$ AUD per sample. We note that only the values for NPK were returned from the commercial laboratory, which is used in the following analysis. Fig. \ref{fig:ExpRealSample} shows an example of a sample obtained by the \Sampler, during the transportation process to the \Analyzer, and additional samples collected for post-analysis by a commercial laboratory.

% \begin{figure}[t]
% \centering
%     \includegraphics[width=\linewidth]{Figures/ExpRealSample.png}
%     \caption{a) A soil sample acquired by the \Sampler. b) A soil sample getting transferred to the \Analyzer. c) Samples collected for ground truth analysis.}
% \label{fig:ExpRealSample}
% \end{figure}

%%%%%%%%%%%%%%%%%%%%%%%%%%%%%%%%%%%%%%%%%%
%============ [ Subsection ] ============%
%%%%%%%%%%%%%%%%%%%%%%%%%%%%%%%%%%%%%%%%%%
\subsection{Evaluation of \Sampler\ and \Analyzer}

% \begin{figure}[t]
% \centering
%     \includegraphics[width=\linewidth]{Figures/AllOperationDataOneColumn.png}
%     \caption{Operational data from the field trial for the \Analyzer\ (a) and  \Sampler\ (b).}
% \label{fig:AllOperationData}
% \end{figure}

\begin{figure}[t]
\captionsetup[subfigure]{justification=centering}
\centering
    \begin{subfigure}{1\columnwidth}
        \centering
        \includegraphics[width=\linewidth]{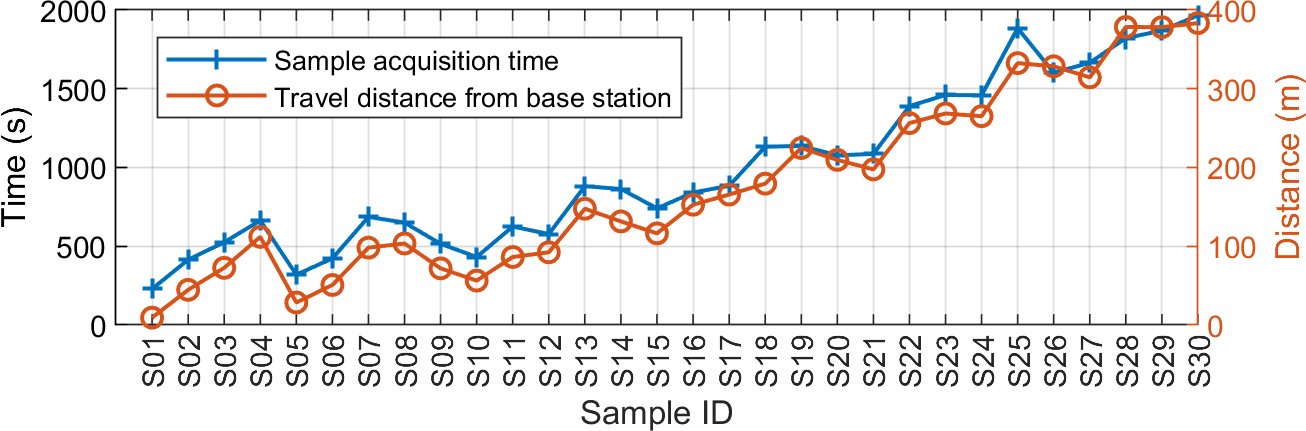}
        \caption{Operational data for the \Sampler}
        \label{fig:AllOperationData_a}
    \end{subfigure}
    \begin{subfigure}{1\columnwidth}
        \centering
        \includegraphics[width=\linewidth]{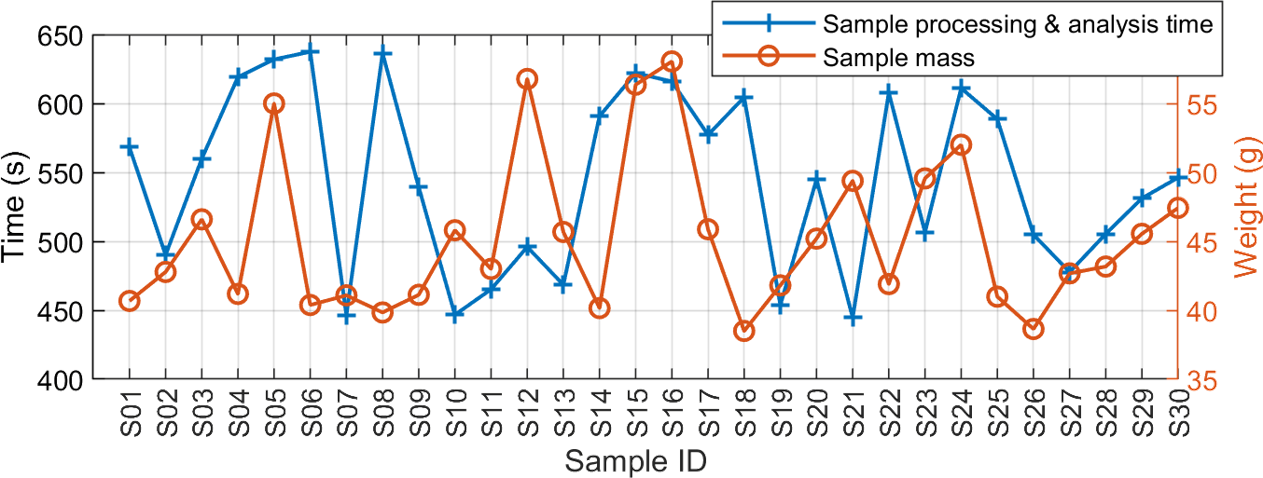}
        \caption{Operational data for the \Analyzer}
        \label{fig:AllOperationData_b}
    \end{subfigure}
    \caption{Operational data from the multi-day field trial.}
    \label{fig:AllOperationData}
\end{figure}

Fig. \ref{fig:AllOperationData} illustrates the operational data recorded during the field trial for both the \Analyzer\ and the \Sampler. For the \Sampler, sample acquisition time is plotted alongside travel distance. It is obvious that greater travel distances result in longer sample acquisition times. Among the samples, S25 exhibits a sudden spike in acquisition time due to challenging terrain. Specifically, the area between S24 and S25 (Fig. \ref{fig:FieldTrialOverview}) featured a steep slope and dense grass, necessitating manual takeover to navigate the robot safely to the sample location.
For the \Analyzer, sample processing and analysis times are compared with sample mass, showing how these factors varied from sample to sample. The sample mass remains relatively stable with most samples weighing between 40 and 50 grams, which is well within the target range of 30–50 grams, and the average weight is 45.2 grams. In contrast, sample processing and analysis times fluctuate, with an average of 544.6 seconds (approximately 9 minutes). These fluctuations may be attributed to unaccounted factors such as sample composition or sensor's long-term performance.% emphasizing the complexity and variability of the \Analyzer's processing steps.

% \begin{figure}[t]
% \centering
%     \includegraphics[width=\linewidth]{Figures/AllISFETDataOneColumn.png}
%     \caption{a) pH measurements and b) percent error for NPK measurements obtained for all samples. The mean measurement errors for N, P and K concentration levels are 7.8\%, 13\% and 7.1\%, respectively.}
% \label{fig:AllISFETData}
% \end{figure}

% \begin{figure}[t]
% \centering
%     \includegraphics[width=\linewidth]{Figures/pHresults.png}
%     \caption{a) Example of pH sensor's calibration points. b) Correlation of soil pH values as measured by our Lab and the commercial pH Probe.}
% \label{fig:pHresults}
% \end{figure}. I 

% The pH measurement accuracy of the \Analyzer\ is evaluated with 10 soil samples of varying pH levels. As can be seen in Fig. \ref{fig:pHresults}, a high correlation between the pH values measured by our \Analyzer\ and a commercial soil pH probe\footnote{\url{https://shorturl.at/5TZ5a}} was observed, with $R^2=0.87$. This result indicates a strong correlation between the ISFET sensor measurements and the pH values from a commercial product. However, there may be some minor inconsistencies due to sensor noise, environmental factors, or measurement errors by the soil pH probe itself.
% Fig. \ref{fig:AllISFETData} presents the measured pH values and the percent errors of NPK for the soil samples from the field trials. 

Using the NPK concentration values reported by the commercial laboratory, which were available 3 weeks after the samples were sent for analysis, 7.8\%, 13\%, and 7.1\% were found to be the mean percent errors for NPK respectively. The relatively low errors for Nitrogen and Potassium suggest that the \Analyzer\ is well-optimized for these macronutrients. However, the higher error observed for Phosphorus could be attributed to the use of a less suitable reagent for Phosphorus, as noted in \cite{riedel2024concept}. These findings demonstrate the \Analyzer's capability for accurate macronutrient analysis while also identifying opportunities for improvement, particularly in the detection of Phosphorus.

%%%%%%%%%%%%%%%%%%%%%%%%%%%%%%%%%%%%%%%%%%
%============ [ Subsection ] ============%
%%%%%%%%%%%%%%%%%%%%%%%%%%%%%%%%%%%%%%%%%%
\subsection{Evaluation of Soil Mapping Accuracy}

The root-mean-square error (RMSE) is used to evaluate the mapping accuracy for each element. The RMSE can demonstrate our system's predictive performance, which is defined as $\text{RMSE}_i = \sqrt{\frac{1}{M} \sum_{j=1}^{M} (\bar{y}_i(\InputVec_j) - y^*_i(\InputVec_j))^2}$ where $i\in[1,3]$ is the index of the element (NPK), $\bar{y}_i$ and $y^*_i$ are the true and predicted values of the $i$-th element, and $M$ is the total number of points in the predicted domain.

% \begin{figure}[t]
% \centering
%     \includegraphics[width=\linewidth]{Figures/CompareSTGPvsMTGP.png}
%     \caption{RMSE results obtained by (a) \PrevLearner\ and (b) \LearnerName\ methods.}
% \label{fig:CompareSTGPvsMTGP}
% \end{figure}

\begin{figure}[t]
\captionsetup[subfigure]{justification=centering}
\centering
    \begin{subfigure}{.49\columnwidth}
        \centering
        \includegraphics[width=\linewidth]{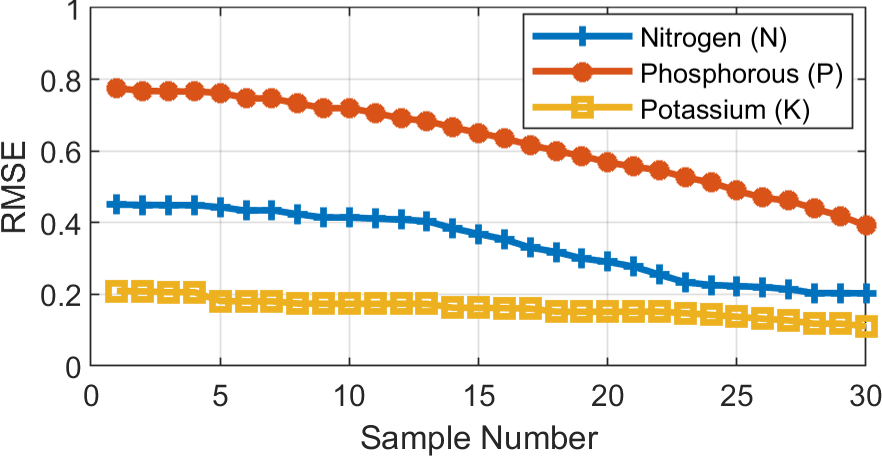}
        \caption{\PrevLearner}
        \label{fig:CompareSTGPvsMTGP_a}
    \end{subfigure}
    \begin{subfigure}{.49\columnwidth}
        \centering
        \includegraphics[width=\linewidth]{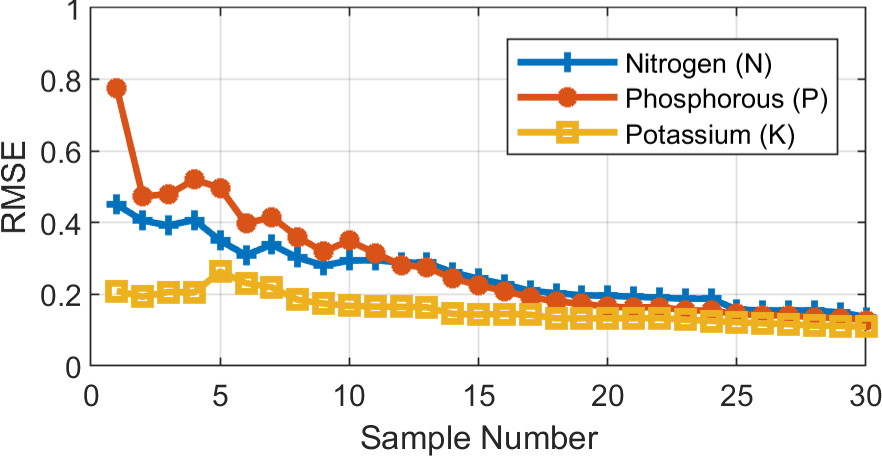}
        \caption{\LearnerName}
        \label{fig:CompareSTGPvsMTGP_b}
    \end{subfigure}
    \caption{Comparison of RMSE between different methods.}
    \label{fig:CompareSTGPvsMTGP}
\end{figure}

% \begin{figure}[t]
% \centering
%     \includegraphics[width=\linewidth]{Figures/AllCorrCoeffNPK.png}
%     \caption{Estimated correlation coefficients for (a) N and (b) P.}
% \label{fig:AllCorrCoeffNPK}
% \end{figure}

\begin{figure}[t]
\captionsetup[subfigure]{justification=centering}
\centering
    \begin{subfigure}{.49\columnwidth}
        \centering
        \includegraphics[width=\linewidth]{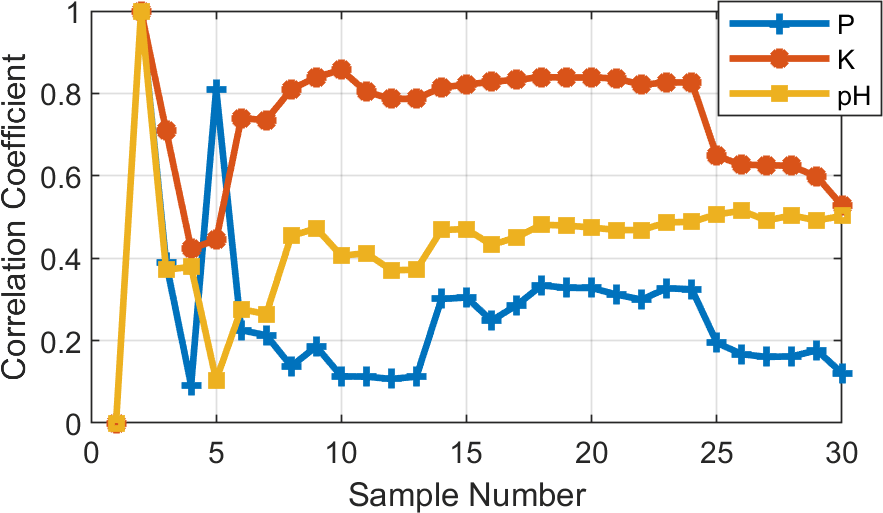}
        \caption{Estimated correlation for N}
        \label{fig:AllCorrCoeffNPK_a}
    \end{subfigure}
    \begin{subfigure}{.49\columnwidth}
        \centering
        \includegraphics[width=\linewidth]{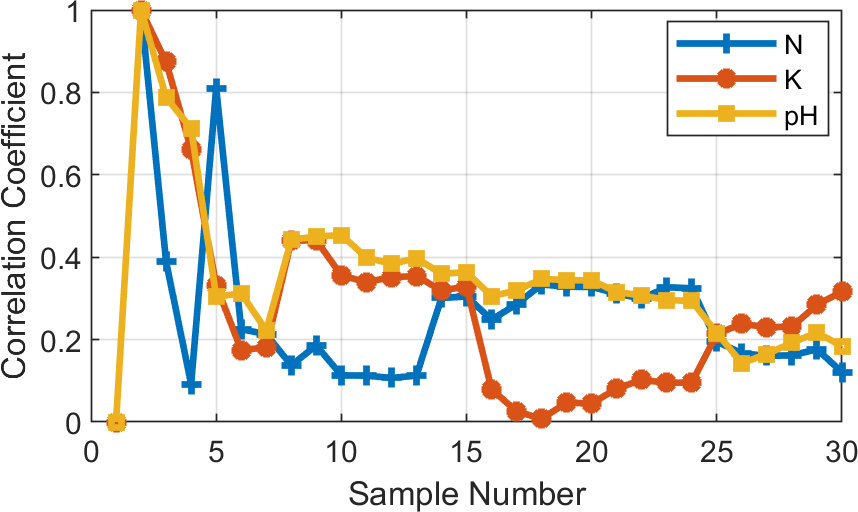}
        \caption{Estimated correlation for P}
        \label{fig:AllCorrCoeffNPK_b}
    \end{subfigure}
    \caption{Estimated correlation as a function of sample size.}
    \label{fig:AllCorrCoeffNPK}
\end{figure}

Fig. \ref{fig:CompareSTGPvsMTGP} compares the RMSE results for NPK analysis across 30 samples using \PrevLearner\ and our \LearnerName\ methods. Both methods process the samples sequentially, from S01 to S30. While the RMSE values for both methods are the same for the first sample, by the last sample the RMSE for \PrevLearner\ reaches up to 0.4, indicating a significantly higher error margin compared to \LearnerName, which achieves a maximum RMSE of approximately 0.14. Additionally, the error reduction rate for \PrevLearner\ is noticeably slower than for \LearnerName, underscoring \LearnerName's advantage in leveraging correlations across elements to enhance prediction accuracy. Fig. \ref{fig:AllCorrCoeffNPK} presents the estimated correlation coefficients between the macronutrients, which cannot be obtained with the \PrevLearner\ method. The results show no significant correlation between the N-P and K-pH pairs (coefficients close to $0$), while the relationships between other pairs remain inconclusive. These findings demonstrate the superior performance of \LearnerName\ over \PrevLearner\ in reducing prediction errors and improving accuracy for soil nutrient mapping tasks, while also offering the added benefit of estimating inter-element correlations.

\begin{figure}[t]
\captionsetup[subfigure]{justification=centering}
\centering
    \begin{subfigure}{1\columnwidth}
        \centering
        \includegraphics[width=\linewidth]{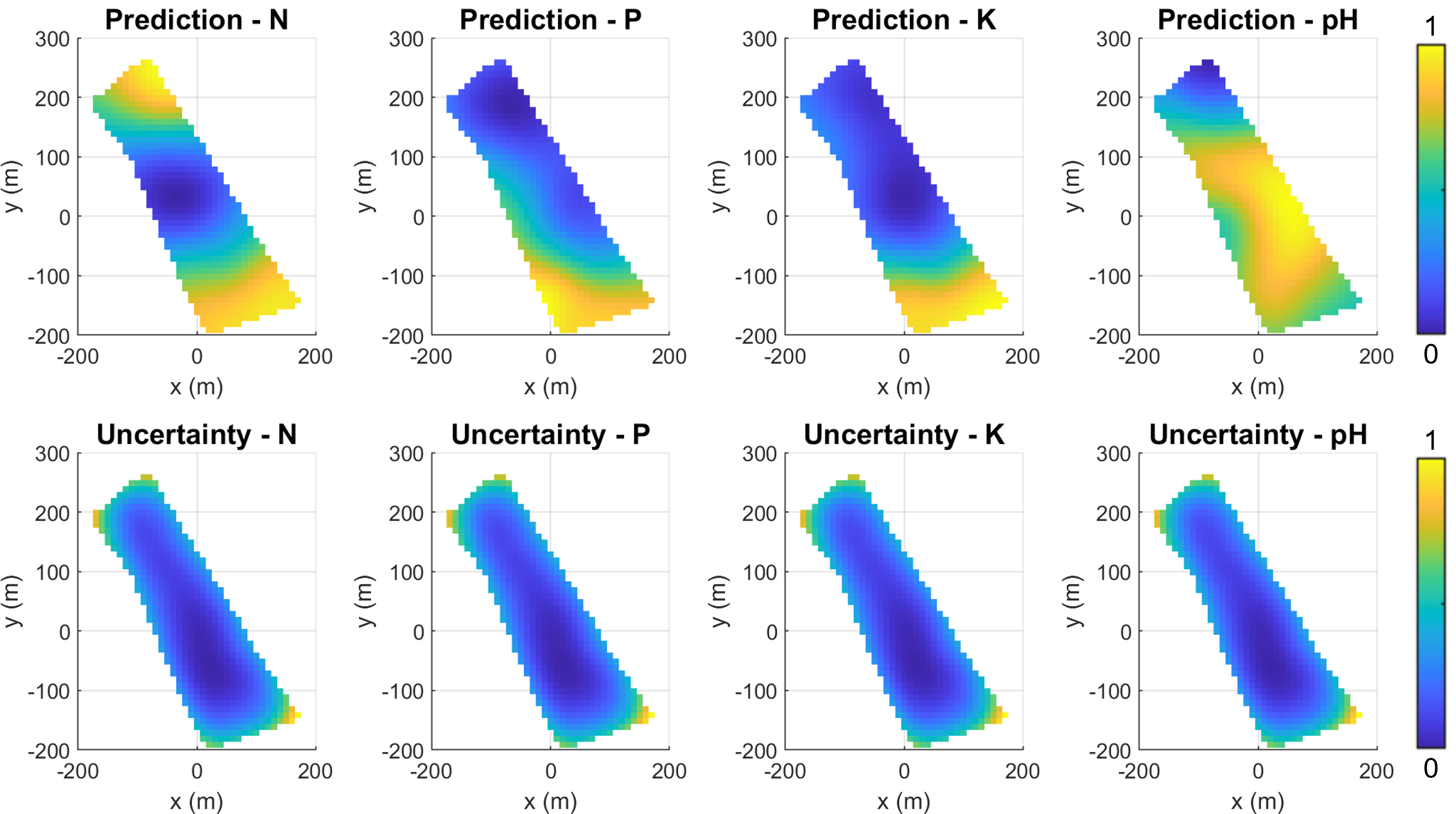}
        \caption{Results obtained by \PrevLearner}
        \label{fig:ResultsRodsFarmMaps_a}
        \vspace{0.4cm}
    \end{subfigure}
    \begin{subfigure}{1\columnwidth}
        \centering
        \includegraphics[width=\linewidth]{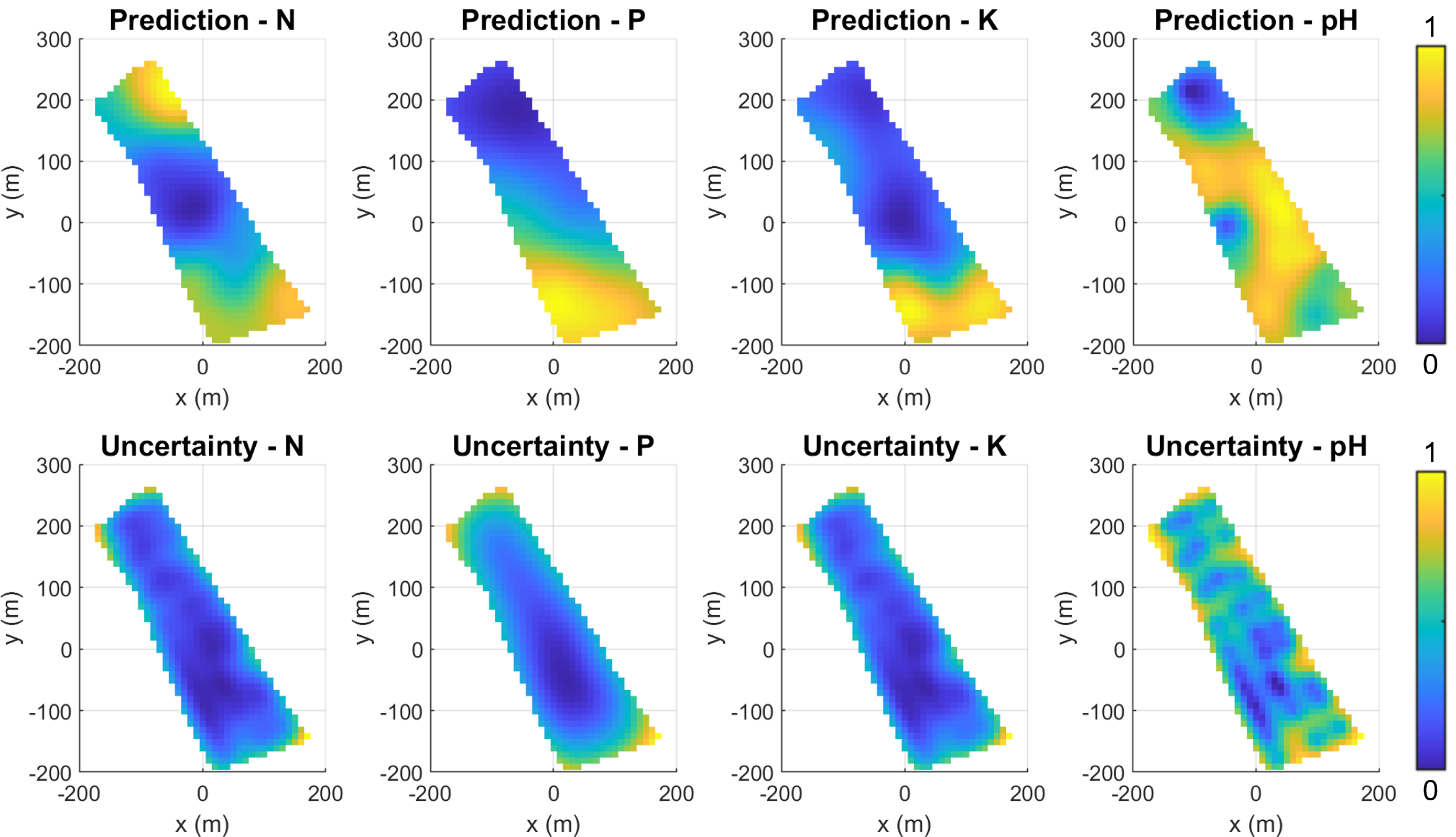}
        \caption{Results obtained by \LearnerName}
        \label{fig:ResultsRodsFarmMaps_b}
    \end{subfigure}
    \caption{Prediction and uncertainty maps for soil NPK and pH generated by different modelling methods. The values within each map have been normalized.}
    \label{fig:ResultsRodsFarmMaps}
\end{figure}

Fig. \ref{fig:ResultsRodsFarmMaps} illustrates the predicted spatial distributions and associated uncertainties of NPK and pH levels in the farm area when all 30 samples are processed. The prediction maps in the top row reveal nutrient availability and soil acidity across the field. Notably, the central and upper regions of the field show relatively low and uniform nutrient levels. Since these are the main grazing areas for cattle, grass is regularly consumed which might explain why nutrients are depleted. In contrast, the lower regions of the maps display higher and more variable nutrient levels, possibly because these areas are less frequented by cattle, allowing for nutrient accumulation. In these lower region, the maps obtained by our \LearnerName\ are much more detailed compared to \PrevLearner, which demonstrates the effectiveness of the proposed method.
Similarly, with the uncertainty maps, while \PrevLearner\ produces uniform uncertainty in most regions, \LearnerName\ was able to leverage the inter-correlation between variables to produce much more nuanced results.
%Regions near the sampled points exhibit low uncertainty, reflecting high confidence in the predictions, while areas farther from the sampling locations show increased uncertainty.
These maps demonstrate how our \LearnerName\ method can generate more accurate and detailed predictions across the entire environment compared to \PrevLearner.

%%%%%%%%%%%%%%%%%%%%%%%%%%%%%%%%%%%%%%%%%%
%                SECTION                 %
%%%%%%%%%%%%%%%%%%%%%%%%%%%%%%%%%%%%%%%%%%
\section{LESSONS LEARNED AND CHALLENGES} \label{sec:Lessons}

% \begin{figure*}[t]
% \centering
%     \includegraphics[width=\textwidth]{Figures/ObstacleChallenges.png}
%     \caption{Challenges for autonomous systems in agricultural environments: detecting obscure obstacles such as (a) wooden logs, (b) wire fences, and (c) sprinklers or water pipes; (d) traversing difficult terrains; (e) operating under adverse weather conditions or in the presence of farm animals.}
%     \label{fig:ObstacleChallenges}
% \end{figure*}

\begin{figure*}[t]
\captionsetup[subfigure]{justification=centering}
\centering
    \begin{subfigure}{.4\columnwidth}
        \centering
        \includegraphics[width=\linewidth]{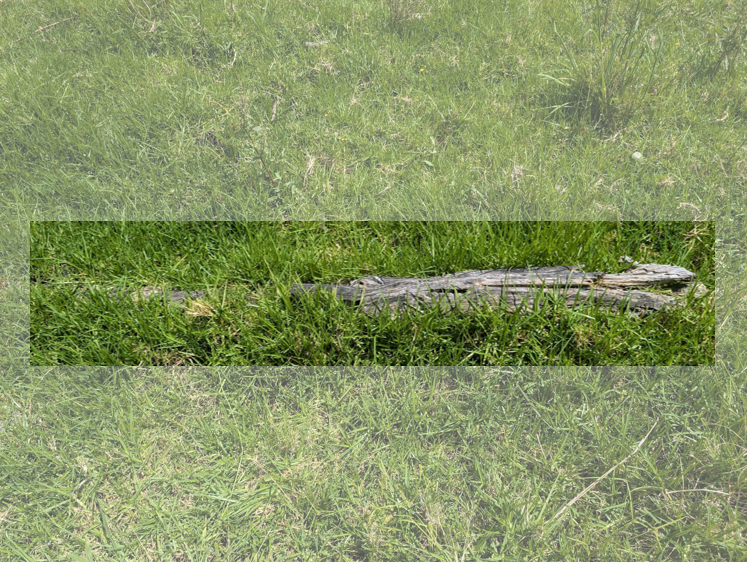}
        \caption{}
        \label{fig:ObstacleChallenges_a}
    \end{subfigure}
    \begin{subfigure}{.4\columnwidth}
        \centering
        \includegraphics[width=\linewidth]{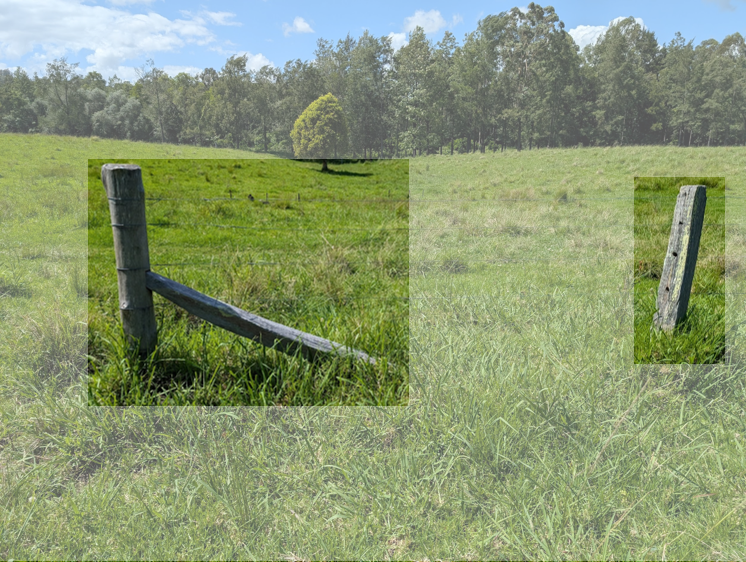}
        \caption{}
        \label{fig:ObstacleChallenges_b}
    \end{subfigure}
    \begin{subfigure}{.4\columnwidth}
        \centering
        \includegraphics[width=\linewidth]{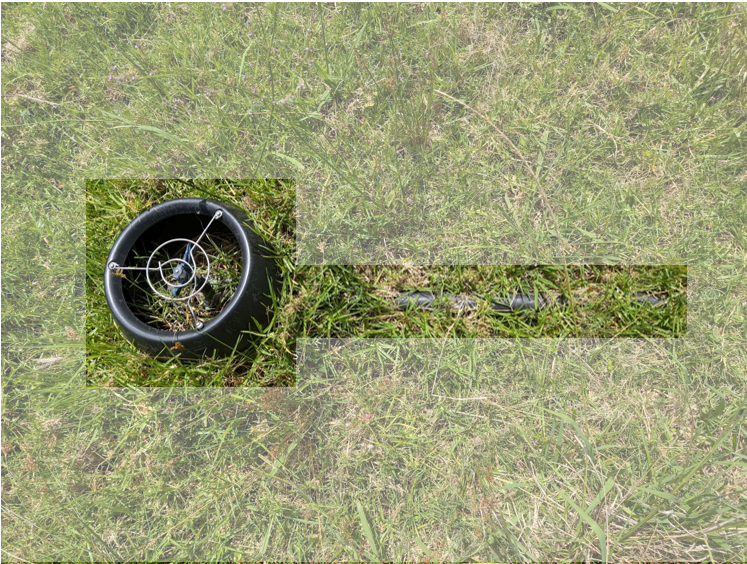}
        \caption{}
        \label{fig:ObstacleChallenges_c}
    \end{subfigure}
    \begin{subfigure}{.4\columnwidth}
        \centering
        \includegraphics[width=\linewidth]{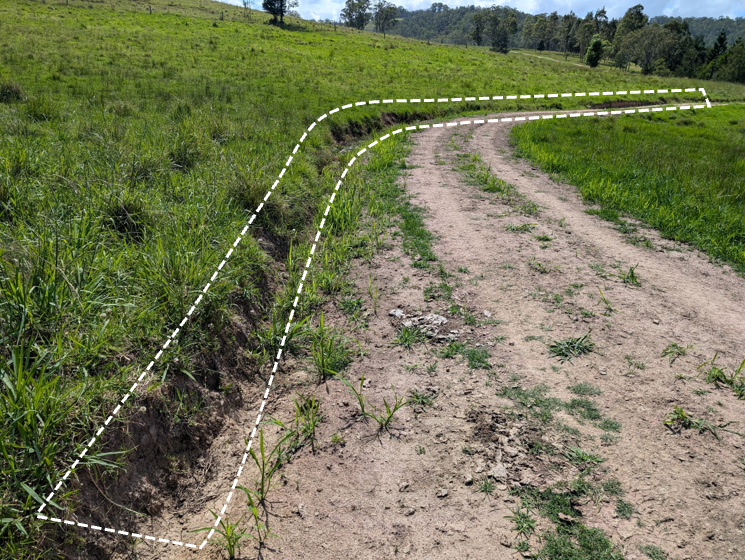}
        \caption{}
        \label{fig:ObstacleChallenges_d}
    \end{subfigure}
    \begin{subfigure}{.4088\columnwidth}
        \centering
        \includegraphics[width=\linewidth]{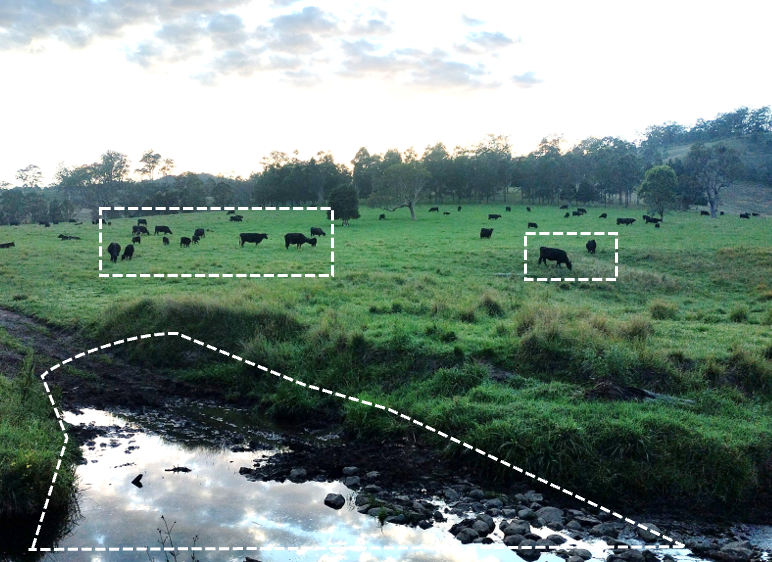}
        \caption{}
        \label{fig:ObstacleChallenges_e}
    \end{subfigure}
    \caption{Challenges for autonomous systems in agricultural environments: detecting obscure obstacles such as (a) wooden logs, (b) wire fences, and (c) sprinklers or water pipes; (d) traversing difficult terrains; (e) operating under adverse weather conditions or in the presence of farm animals.}
    \label{fig:ObstacleChallenges}
\end{figure*}

%%%%%%%%%%%%%%%%%%%%%%%%%%%%%%%%%%%%%%%%%%
%============ [ Subsection ] ============%
%%%%%%%%%%%%%%%%%%%%%%%%%%%%%%%%%%%%%%%%%%
\subsection{Mechatronic Complexities}

For the \Sampler, a major challenge was selecting a soil acquisition mechanism that could be integrated with the DFH robot while minimizing mechanical complexity. Two prototype samplers were developed and tested: one using a probe and the other an auger. As the name suggests, the probe sampler relies on a soil probe for ground penetration and sample extraction, whereas the auger sampler uses a rotating auger for the same purpose. Mechanically, the auger sampler is simpler than the probe sampler, since the former only needs to be held in place and does not generate significant upward forces during sample acquisition, which means it does not require dedicated supports. As such, the auger sampler was selected as the final design. The \Sampler\ was then developed and continuously tested with different soil types, moisture levels, and hardness levels to ensure its reliability and robustness. Nevertheless, the inherent variability in soil composition within a single farm and across Australia's diverse landscape introduces substantial mechanical challenges. Developing a robotic solution capable of handling various soil conditions, from sandy to clay-rich, from wet to compacted soil, is vital for creating a practical autonomous soil sampling system for real-world applications \cite{carter2007soil}.

For the \Analyzer, the main challenge lies in coordinating multiple physical operations, particularly the design and implementation of the sample processing pipeline. Significant effort was dedicated to ensuring that the pipeline: 1) matches closely with an established workflow for ISFET-based soil nutrient analysis \cite{najdenko2023development}, 2) functions seamlessly in the field - from fluid pumping, auger mixing, centrifuging, to sensor conditioning and protection. Despite successful demonstrations in real farm conditions, several major issues remain. First, once a sample is fully processed, the sample cuvette might be clogged due to plant residue and small stones, which remains a critical obstacle in developing the self-cleaning capability for the \Analyzer. Additionally, maintaining the accuracy of the ISFET sensor still requires a number of manual steps, from conditioning and calibrating the sensors to dry storage. Failure to perform these steps often resulted in unacceptable noise levels in the sensor data, compromising measurement accuracy or even the sensors themselves. Overcoming these challenges is crucial for developing a fully autonomous system.

%%%%%%%%%%%%%%%%%%%%%%%%%%%%%%%%%%%%%%%%%%
%============ [ Subsection ] ============%
%%%%%%%%%%%%%%%%%%%%%%%%%%%%%%%%%%%%%%%%%%
\subsection{Environmental Complexities}

From a navigation and control perspective, designing a robot to follow a GPS-guided path, collect samples, and return to base may seem straightforward. However, real farm environments present a range of unforeseen challenges, as illustrated in Fig. \ref{fig:ObstacleChallenges}. First, the obstacle detection algorithm might fail to distinguish obscure objects from their surrounding (Fig. \ref{fig:ObstacleChallenges_a}-\ref{fig:ObstacleChallenges_b}). Second, the robot needs to operate carefully around infrastructure such as sprinklers and water pipes (Fig. \ref{fig:ObstacleChallenges_c}) to avoid damaging them, without prior knowledge of their exact locations on the field. Third, uneven or weather-affected terrains (Fig. \ref{fig:ObstacleChallenges_d}-\ref{fig:ObstacleChallenges_e}) can hinder the robot’s locomotion and stability. Last but not least, the robot will need to coexist with farm animals that may share its operating space (Fig. \ref{fig:ObstacleChallenges_e}), a challenge that traditional navigation algorithms have not been specifically designed to handle. Managing these real-world variations, in particular changes in terrain, weather conditions, unknown obstacles, and the presence of farm animals, is essential to improve the overall performance and reliability of the system.

%%%%%%%%%%%%%%%%%%%%%%%%%%%%%%%%%%%%%%%%%%
%============ [ Subsection ] ============%
%%%%%%%%%%%%%%%%%%%%%%%%%%%%%%%%%%%%%%%%%%
\subsection{Sampling Strategy Efficiency}

Our current system follows a uniform sampling strategy, but not all sampling points contribute equally to improving prediction accuracy. However, as evidenced by this paper, extracting and processing each soil sample are complex and costly processes and should only be conducted when necessary. As shown in Fig. \ref{fig:CompareSTGPvsMTGP_b}, the RMSEs for all elements show minimal reduction from sample 17 to 24, suggesting that some of these samples could potentially be skipped to save time and resources.
Hence, an adaptive sampling strategy can be employed to prioritize the most informative samples based on prior data as well as the inter-task correlations \cite{Nick2024ICRA}. However, a key challenge with our system is the inherent delay between sample acquisition and obtaining measurement results. Hence, we aim to develop an adaptive sampling strategy that leverages the prediction and uncertainty estimates from \LearnerName\ while accounting for the delay in information gain, which will further enhance the scalability and efficiency of the proposed system.

%%%%%%%%%%%%%%%%%%%%%%%%%%%%%%%%%%%%%%%%%%
%============ [ Subsection ] ============%
%%%%%%%%%%%%%%%%%%%%%%%%%%%%%%%%%%%%%%%%%%
\subsection{Importance of Interdisciplinary Collaboration}

Interdisciplinary collaboration played a crucial role in the development and refinement of our robotic soil sampling and analysis system. Inputs from soil scientists were essential in defining the scientific requirements for the \Sampler\ and \Analyzer. Their expertise ensured that selected sensors met the necessary accuracy and sensitivity requirements for detecting the targeted soil properties, and the sample processing pipeline is aligned with established laboratory protocols. Additionally, discussions with farmers provided valuable insights into real-world operational constraints, such as the need for minimal soil disturbance, ease of maintenance, and compatibility with existing farm workflows. Moving forward, we aim to integrate these different perspectives to develop a system that is not only scientifically sound and technically robust but also practical for real-world agricultural applications.

%%%%%%%%%%%%%%%%%%%%%%%%%%%%%%%%%%%%%%%%%%
%              CONCLUSIONS               %
%%%%%%%%%%%%%%%%%%%%%%%%%%%%%%%%%%%%%%%%%%
\section{CONCLUSIONS}

In this work, we present a modular robotic system for real-time sampling and analysis of soil properties. Our system consists of two key sub-systems: the \Sampler, which is responsible for automated and reliable soil sampling at precise locations and depths; and the \Analyzer, which can measure soil pH levels and NPK concentrations following a semi-automated workflow. 
Extensive multi-day field trials on a large-scale pasture farm validated all sub-systems’ accuracy and performance across all stages, including soil sample acquisition, sample processing, and sample analysis. The results show that our system can obtain accurate spatially continuous maps of key soil properties along with their correlation coefficients.
Overall, the \Sampler\ is capable of acquiring soil samples at 200mm depth with a mass of 40-50g. Within 10 minutes, the \Analyzer\ can process a sample and measure pH with high reliability ($R^2 = 0.87$) as well as NPK ion concentrations with mean errors of 7.8\% (N), 13\% (P) and 7.1\% (K). 

During development, the most challenging engineering aspects of the proposed system lie in the mechatronic complexities of integrating various sub-systems while ensuring that they work well together and perform reliably under real farm conditions.
Additionally, the unstructured nature of the agricultural environments can introduce significant challenges for current perception, navigation and control algorithms.
To achieve a fully autonomous robotic soil sampling and analysis system, our future efforts will focus on boosting the \Sampler's efficiency with a suitable adaptive sampling method, enhancing the \Analyzer's autonomy with self-cleaning and self-calibration mechanisms, and developing more robust perception algorithms to tackle the dynamic and unstructured nature of large-scale agriculture environments.

%%%%%%%%%%%%%%%%%%%%%%%%%%%%%%%%%%%%%%%%%%
%             SPECIAL COMMAND            %
%%%%%%%%%%%%%%%%%%%%%%%%%%%%%%%%%%%%%%%%%%
% \addtolength{\textheight}{-12cm}   % This command serves to balance the column lengths on the last page of the document manually. It shortens the textheight of the last page by a suitable amount. This command does not take effect until the next page so it should come on the page before the last. Make sure that you do not shorten the textheight too much.

%%%%%%%%%%%%%%%%%%%%%%%%%%%%%%%%%%%%%%%%%%
%                SECTION                 %
%%%%%%%%%%%%%%%%%%%%%%%%%%%%%%%%%%%%%%%%%%
\section*{ACKNOWLEDGMENT}

The authors would like to thank the Agriculture and Environment group at ACFR for their continued support in developing, designing, and testing the system.% 

%%%%%%%%%%%%%%%%%%%%%%%%%%%%%%%%%%%%%%%%%%%%%%%%%%%%%%%%%%%%%%%
%% Print bibliography
%%%%%%%%%%%%%%%%%%%%%%%%%%%%%%%%%%%%%%%%%%%%%%%%%%%%%%%%%%%%%%%
\bibliographystyle{IEEEtran}
\bibliography{IEEEabrv,./references}

\begin{thebibliography}{10}
\providecommand{\url}[1]{#1}
\csname url@samestyle\endcsname
\providecommand{\newblock}{\relax}
\providecommand{\bibinfo}[2]{#2}
\providecommand{\BIBentrySTDinterwordspacing}{\spaceskip=0pt\relax}
\providecommand{\BIBentryALTinterwordstretchfactor}{4}
\providecommand{\BIBentryALTinterwordspacing}{\spaceskip=\fontdimen2\font plus
\BIBentryALTinterwordstretchfactor\fontdimen3\font minus \fontdimen4\font\relax}
\providecommand{\BIBforeignlanguage}[2]{{%
\expandafter\ifx\csname l@#1\endcsname\relax
\typeout{** WARNING: IEEEtran.bst: No hyphenation pattern has been}%
\typeout{** loaded for the language `#1'. Using the pattern for}%
\typeout{** the default language instead.}%
\else
\language=\csname l@#1\endcsname
\fi
#2}}
\providecommand{\BIBdecl}{\relax}
\BIBdecl

\bibitem{mcfadden2023precision}
J.~McFadden, E.~Njuki, and T.~Griffin, ``{Precision Agriculture in the Digital Era: Recent Adoption on U.S. Farms},'' \emph{U.S. Department of Agriculture, Economic Research Service}, 2023.

\bibitem{dattatreya2023conventional}
S.~Dattatreya, A.~N. Khan, K.~Jena, and G.~Chatterjee, ``{Conventional to Modern Methods of Soil NPK Sensing: A Review},'' \emph{IEEE Sensors Journal}, 2023.

\bibitem{rayment2011soil}
G.~E. Rayment and D.~J. Lyons, \emph{Soil Chemical Methods - Australasia}.\hskip 1em plus 0.5em minus 0.4em\relax CSIRO Publishing, 2010.

\bibitem{Nick2024ICRA}
N.~Harrison, N.~Wallace, and S.~Sukkarieh, ``{Automated Testing of Spatially-Dependent Environmental Hypotheses through Active Transfer Learning},'' in \emph{{2024 IEEE International Conference on Robotics and Automation (ICRA)}}, 2024.

\bibitem{botta2022review}
A.~Botta, P.~Cavallone, L.~Baglieri, G.~Colucci, L.~Tagliavini, and G.~Quaglia, ``A review of robots, perception, and tasks in precision agriculture,'' \emph{applied mechanics}, vol.~3, no.~3, pp. 830--854, 2022.

\bibitem{harun2023robotic}
H.~Harun, ``A robotic system for in-situ measurement of soil total carbon and nitrogen,'' Master's Thesis, University of Nebraska - Lincoln, Lincoln, Nebraska, July 2023, advisor: Yufeng Ge.

\bibitem{sibley2008development}
K.~J. Sibley, \emph{Development and use of an automated on-the-go soil nitrate mapping system}.\hskip 1em plus 0.5em minus 0.4em\relax Wageningen University and Research, 2008.

\bibitem{kitic2022agrobot}
G.~Kiti{\'c}, D.~Krklje{\v{s}}, M.~Pani{\'c}, C.~Petes, S.~Birgermajer, and V.~Crnojevi{\'c}, ``{Agrobot Lala—an autonomous robotic system for real-time, in-field soil sampling, and analysis of nitrates},'' \emph{Sensors}, vol.~22, no.~11, p. 4207, 2022.

\bibitem{hinck2022prototypes4soil2data}
S.~Hinck, V.~Riedel, A.~Ruckelshausen, A.~M{\"o}ller, M.~Terhaag, T.~Meyer, D.~Mentrup, H.~Kerssen, E.~Najdenko, F.~Lorenz \emph{et~al.}, ``Analysis-to-go on the field: prototypes4soil2data,'' in \emph{Proceedings of the 22nd World Congress of Soil Science}, Glasgow, Scotland, 2022.

\bibitem{ruckelshausen2009bonirob}
A.~Ruckelshausen, P.~Biber, M.~Dorna, H.~Gremmes, R.~Klose, A.~Linz, R.~Rahe, R.~Resch, M.~Thiel \emph{et~al.}, ``Bonirob: an autonomous field robot platform for individual plant phenotyping,'' in \emph{Precision agriculture'09}.\hskip 1em plus 0.5em minus 0.4em\relax Wageningen Academic, 2009, pp. 841--847.

\bibitem{grotzinger2012mars}
J.~P. Grotzinger, J.~Crisp, A.~R. Vasavada, R.~C. Anderson, C.~J. Baker, R.~Barry, D.~F. Blake, P.~Conrad, K.~S. Edgett, B.~Ferdowski \emph{et~al.}, ``Mars science laboratory mission and science investigation,'' \emph{Space science reviews}, vol. 170, pp. 5--56, 2012.

\bibitem{riedel2024concept}
V.~Riedel, S.~Hinck, E.~Peiter, and A.~Ruckelshausen, ``{Concept and Realisation of ISFET-Based Measurement Modules for Infield Soil Nutrient Analysis and Hydroponic Systems},'' \emph{Electronics}, vol.~13, no.~13, p. 2449, 2024.

\bibitem{najdenko2023development}
E.~Najdenko, F.~Lorenz, H.-W. Olfs, and K.~Dittert, ``Development of an express method for measuring soil nitrate, phosphate, potassium, and ph for future in-field application,'' \emph{Journal of Plant Nutrition and Soil Science}, vol. 186, no.~6, pp. 623--632, 2023.

\bibitem{williams2006gaussian}
C.~K. Williams and C.~E. Rasmussen, \emph{Gaussian processes for machine learning}.\hskip 1em plus 0.5em minus 0.4em\relax MIT press Cambridge, MA, 2006, vol.~2.

\bibitem{bonilla2007multi}
E.~V. Bonilla, K.~Chai, and C.~Williams, ``Multi-task gaussian process prediction,'' \emph{Advances in neural information processing systems}, vol.~20, 2007.

\bibitem{durichen2014multitask}
R.~D{\"u}richen, M.~A. Pimentel, L.~Clifton, A.~Schweikard, and D.~A. Clifton, ``Multitask gaussian processes for multivariate physiological time-series analysis,'' \emph{IEEE Transactions on Biomedical Engineering}, vol.~62, no.~1, pp. 314--322, 2014.

\bibitem{melkumyan2011multi}
A.~Melkumyan and F.~Ramos, ``{Multi-kernel Gaussian processes},'' in \emph{IJCAI Proceedings-International Joint Conference on Artificial Intelligence}, vol.~22, no.~1, 2011, p. 1408.

\bibitem{carter2007soil}
M.~R. Carter and E.~G. Gregorich, \emph{Soil sampling and methods of analysis}.\hskip 1em plus 0.5em minus 0.4em\relax CRC press, 2007.

\end{thebibliography}

\end{document}